\documentclass[]{IEEEtran}
\ifCLASSINFOpdf
  \usepackage[pdftex]{graphicx}
  \graphicspath{{./img/}{.}}
\else
\fi
\usepackage[cmex10]{amsmath}
\AtBeginDocument{%
 \abovedisplayskip=0pt plus 3pt minus 9pt
 \abovedisplayshortskip=0pt plus 3pt
}

\ifCLASSOPTIONcompsoc
  \usepackage[caption=false,font=normalsize,labelfont=sf,textfont=sf]{subfig}
\else
  \usepackage[caption=false,font=footnotesize]{subfig}
\fi
\usepackage{acronym}

\acrodef{STDP}[STDP]{Spike-Timing Dependent Plasticity}
\acrodef{SNN}[SNN]{Spiking Neural Networks}
\acrodef{MPDP}[MPDP]{Membrane Potential Dependent Plasticity}

\begin{document}

\makeatletter





\let\origsection\subsection
\renewcommand\subsection{\@ifstar{\starsubsection}{\nostarsubsection}}

\newcommand\nostarsubsection[1]
{\subsectionprelude\origsection{#1}\subsectionpostlude}

\newcommand\starsbsection[1]
{\subsectionprelude\origsection*{#1}\subsectionpostlude}

\newcommand\subsectionprelude{%
  \vspace{-0.5em}
}

\newcommand\subsectionpostlude{%
  \vspace{0em}

}
\makeatother

%
\title{Membrane-Dependent Neuromorphic Learning Rule for Unsupervised Spike Pattern Detection}

%
%
\author{
 \IEEEauthorblockN{}
 \IEEEauthorblockA{
 Sadique Sheik$^{1}$, Somnath Paul$^2$, Charles Augustine$^2$, Gert Cauwenberghs$^{1,3}$\\
 $^{1}$BioCircuits Institute, UC San Diego, La Jolla, CA , USA\\
 $^{2}$Circuit Research Lab, Intel Corporation, Hillsboro, OR, USA\\
 $^{3}$Department of Bioengineering, UC San Diego, La Jolla, CA , USA\\
 Email: ssheik@ucsd.edu}
}

\maketitle

\pagenumbering{gobble}

\begin{abstract}
Several learning rules for synaptic plasticity, that depend on either spike timing or internal state variables, have been proposed in the past imparting varying computational capabilities to \acp{SNN}.
Due to design complications these learning rules are typically not implemented on neuromorphic devices leaving the devices to be only capable of inference.
In this work we propose a unidirectional post-synaptic potential dependent learning rule that is only triggered by pre-synaptic spikes, and easy to implement on hardware. We demonstrate that such a learning rule is functionally capable of replicating computational capabilities of pairwise STDP. Further more, we demonstrate that this learning rule can be used to learn and classify spatio-temporal spike patterns in an unsupervised manner using individual neurons. We argue that this learning rule is computationally powerful and also ideal for hardware implementations due to its unidirectional memory access.
\end{abstract}


%
\IEEEpeerreviewmaketitle

\section{Introduction}

Neuromorphic devices aim to capitalize on event based temporal processing and unsupervised learning features of brain inspired spiking neural networks, for efficient, low-power and robust computation.
Learning capability in neuromorphic devices is obtained through the use of plastic synapses. Several neuromorphic systems are being developed with the capability to learn synaptic weights of spiking neurons implemented on silicon~\cite{Mayr_etal16}.

One of the major blocks constituting neuromorphic devices with plastic synapses is the memory required to store synaptic weights in addition to the circuitry required to implement the learning rule.
While some neuromorphic devices have dedicated plastic synaptic circuits~\cite{Mitra_etal2006,Qiao_etal2015} or have a cross bar type synaptic memory~\cite{Seo_etal11}, in most neuromorphic devices, synaptic connectivity and weight memory is implemented on external or internal dedicated digital RAM. Implementation of dedicated plastic synapse circuit for every individual synapse dramatically increases the area requirement for neuromorphic chips, which is not desirable. 

A second factor that discourages implementation of plasticity on neuromorphic devices is the added complexity in the type of memory required to implement current spike based learning rules.
Pairwise \ac{STDP}~\cite{Bi_Poo1998} is a widely used learning rule in computational models of spiking neural networks~\cite{Masquelier_etal2008,Nessler_etal2009,Neftci_etal14}. Event driven implementation of this learning rule triggers an update in the synaptic weight both at the arrival of a presynaptic spike and generation of a postsynaptic spike. This necessitates the accesses to memory associated with the corresponding synaptic weight location in a bidirectional manner i.e. both based on the identity both pre-synaptic and postsynaptic neurons.

These factors hold true also for triplet-based \ac{STDP} rule~\cite{Pfister_Gerstner2006}. A physical implementation of such synaptic weight memory therefore has to meet these requirements in order to achieve efficient \ac{STDP}. Traditional RAM devices are not designed for such access in an efficient manner, especially for sparsely connected networks. Instead one requires design of custom memory architecture such as the bidirectional crossbar memory architecture~\cite{Seo_etal11}. While this is a viable solution for the implementation of \ac{STDP} based learning rules, it is very expensive in terms of total memory and silicon area.

Alternate learning rules that rely on the presynaptic spike timing and neuronal state variables are being explored for neuromorphic implementations~\cite{Feldman_2012}. These learning rules also tend to be biophysically plausible compared to the pairwise \ac{STDP}.
The authors of \cite{Brader_etal2007} present a model of spike-driven synaptic plasticity for bistable synapses where synapses are modified on the arrival of presynaptic spikes based on the post-synaptic potential and calcium concentration. They show that such a rule can enable a neuron to learn to classify complex stimuli in a semi-supervised fashion based on the input firing rates.
A calcium voltage dependent plasticity model~\cite{Graupner_Brunel2012} has been shown to approximates \emph{several} \ac{STDP} curves experimentally observed in biology. 
Similarly, a 
\ac{MPDP} rule proposed recently~\cite{Albers_etal2016} has been shown to approximate anti-hebbian \ac{STDP} like learning rule which has been argued to be in agreement with \ac{STDP} of inhibitory synapses~\cite{Haas_etal2006}. The authors show that this model can be used to train a neuron in a supervised manner to spike at precise times.

In this paper, we propose a simple post-synaptic membrane potential dependent event-driven learning rule that statistically emulates pairwise \ac{STDP} like behavior for excitatory synapses.
This learning rule is extremely hardware efficient due to its dependence only on the presynaptic spike timing and not on the postsynaptic spike timing.
In addition we demonstrate that such a learning rule enables a neuron to learn and identify complex temporal spike patterns embedded in a stream of spikes in an unsupervised manner.

\section{Materials and Methods}
A conductance based integrate and fire neuron model has been used to obtain results presented in this paper. The neuron dynamics are described by Eqns. \ref{eqn:lif} and \ref{eqn:lifg}.

\begin{eqnarray}
C_mdV/dt &=& (V_{rest}-V)g_l + (Es-V)g_e 
\label{eqn:lif} \\
\tau_{s}dg_e/dt &=& -g_e + \Sigma_i W_i S(t-t_i) 
\label{eqn:lifg}
\end{eqnarray}
where $V$ is the neuron's membrane potential, $C_m$ is the membrane capacitance, $V_{rest}$ is its resting potential, $E_s$ is the synaptic reversal potential, $g_l$ is the leakage conductance, $g_e$ is the excitatory synaptic conductance, $S$ are the presynaptic spike trains, $W$ is the synaptic weight and $\tau_s$ is the synaptic time constant. 
In addition to the neuronal dynamics, each neuron has a calcium concentration trace associated to it. Calcium dynamics is given by Eqn.~\ref{eqn:calcium}

\begin{equation}
\tau_{ca}dCa/dt = -Ca + S(t-t_{post})
\label{eqn:calcium}
\end{equation}
where $Ca$ is the calcium concentration, $\tau_{ca}$ is the decay time constant and $t_{post}$ is spike timing of post synaptic neuron. The calcium concentration effectively follows a low pass filtered version of neuron's spiking activity.

\subsection{Learning rule}
Each neuron receives inputs from several presynaptic neurons with random initial weights. Over the course of input spike presentation the synaptic weights get updated. The weight update rule consists of two components $\Delta W_v$ and $\Delta W_h$.

\begin{equation}
\Delta W = \Delta W_v + \Delta W_h
\end{equation}

The weight update component based on the post-synaptic potential $\Delta W_{v}$ is given by:

\begin{multline}
	\Delta W_{v} = [ \delta(V_m(t+1) > V_{lth})\eta_+ \\
                    - \delta(V_m(t+1) < V_{lth})\eta_- ] S(t-t_{pre})
\end{multline}
where $\delta(True) = 1$ and $\delta(False) = 0$. $\eta_+$ and $\eta_-$ are the magnitude of positive and negative weight updates. $V_{lth}$ is the membrane threshold voltage that determines whether the weight should be potentiated (LTP) or depressed (LTD). $S$ represents a pre-synaptic spike train.

Homeostatic weight update component $\Delta W_h$ is given by:

\begin{equation}
	\Delta W_{h} = \eta_h(Ca_t-Ca)S(t-t_{pre})
\end{equation}
where $Ca_t$ is the target calcium concentration, $Ca$ is the current calcium concentration and $\eta_h$ is the magnitude of rate of homeostasis.

\subsection{Stability conditions for the learning rule}
In order to ensure that the weights learned are reflective of the statistics of the inputs and do not drift to maximum or minimum bounds together, the parameters of the learning rule need to be appropriately chosen. We postulate the following three criteria:
\textit{i)} For a random spike train at a synapse, with no correlation to the input spike pattern, the weights should drift towards $0$. This is described by the Eqn.~\ref{eqn:weight_drift_negative}.
\textit{ii)} The negative weight updates triggered by spikes uncorrelated with the target spike pattern (when the pattern is not being presented) should not be large enough to nullify the potentiation during the presentation of the pattern. This is described by the Eqn.~\ref{eqn:retention}.
\textit{iii)} The homeostasis should be strong enough to drive the weights up when the firing rate is too low. This is ensured by Eqn.~\ref{eqn:homeostasis_cond}.

\begin{eqnarray}
t_p\eta_+ &\le& t_n\eta_- \label{eqn:weight_drift_negative} \\
\eta_+ &>& t_n f \eta_- \label{eqn:retention} \\
\eta_- &\le& Ca_t \eta_h \label{eqn:homeostasis_cond}
\end{eqnarray}
where $t_p$ is the total duration of the spike pattern, $t_n$ is the average duration of noise or uncorrelated spike patterns presented between patterns.

\subsection{Hardware implementation}

\begin{figure}
\centering
\includegraphics[width=0.5\textwidth]{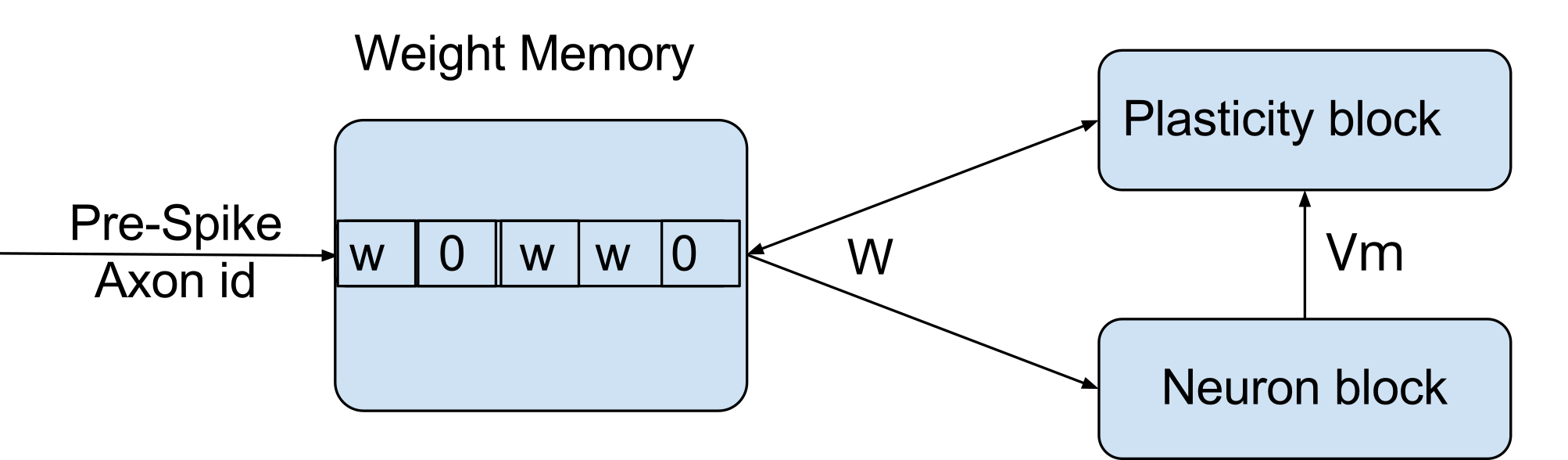}
\caption{A block diagram of data flow in conjunction with neuromorphic hardware using an index/pointer based weight memory.}
\label{fig:hw_memory}
\end{figure}

As previously discussed, synaptic memory constitutes a large fraction of neuromorphic device area, and therefore it is desirable to use techniques that allow compression of this synaptic memory.
Compressed memory schemes such as index based or linked list based routing and weight storage schemes are ideal if the connectivity of the network is sparse. The key limitation of such storage schemes is that of unidirectional memory access ie. if we store connectivity based on source addresses, we can only determine all destinations from a given source (in O(1)) but not all sources given destination. This limits the implementation of learning rules that trigger a weight update by both pre- and post synaptic neurons.
The learning rule we propose here can take advantage of such compressed memory schemes and still perform synaptic plasticity. Fig.~\ref{fig:hw_memory} shows a block diagram of the use of a unidirectional pointer based weight table with the proposed learning rule.

\section{Results}
In order to evaluate the proposed learning rule we conduct a series of experiments described in the following section. Input and output sizes are arbitrarily chosen and verified to work with different sizes (not shown).

\subsection{Equivalence to STDP}
\begin{figure}[!t]
\centering
\includegraphics[width=0.5\textwidth]{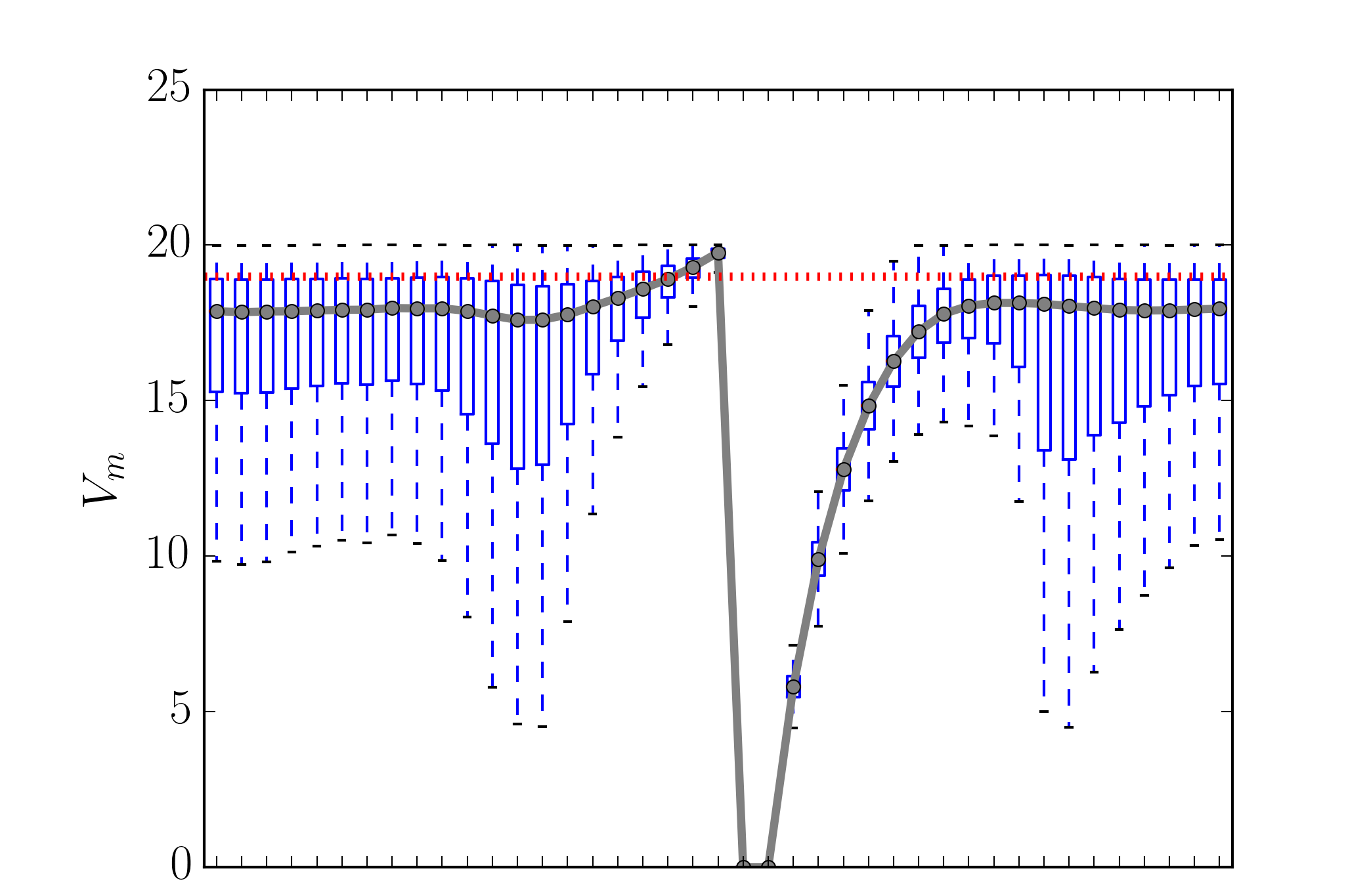}\\
\includegraphics[width=0.5\textwidth]{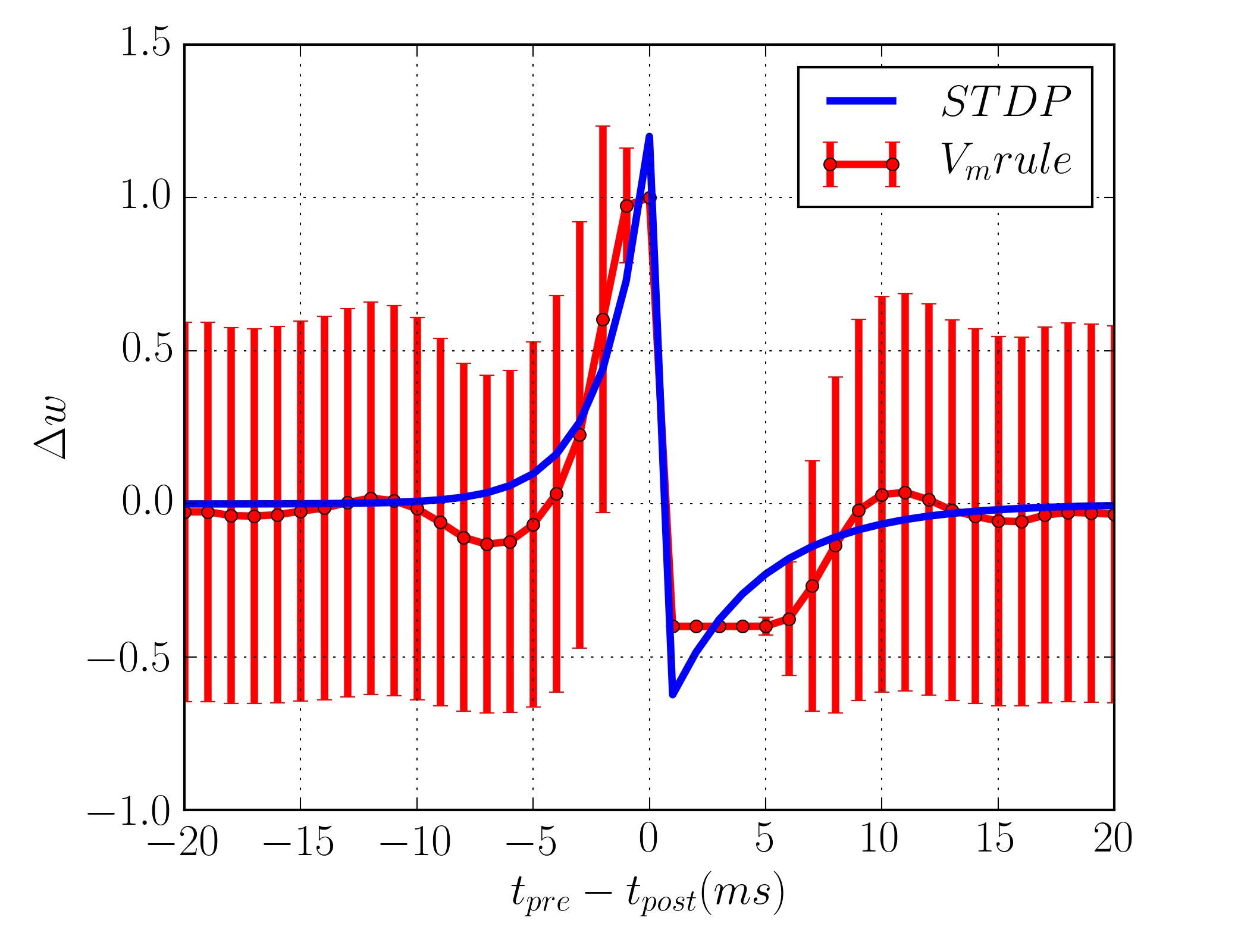}
\caption{(Top) The membrane potential distribution before and after a post-synaptic spike when activated using a stream of Poisson spike trains. The dotted horizontal line in red shows the learning threshold. (Bottom) The mean and standard deviation of weight update before and after a post-synaptic spike as a result of the $V_m$ based learning rule proposed in this paper in red. An empirically equivalent curve of exponential \ac{STDP} rule is plotted in blue.}
\label{fig:vmem_stdp}
\end{figure}

The first experiment consists of a statistical measure of the weight updates triggered by the arrival of a presynaptic spike with respect to postsynaptic spike time. A single post synaptic neuron was presented with random Poisson spike patterns across $225$ synapses. While freezing all the weights, the hypothetical weight updates governed by the learning rule were recorded over time. The recorded data was sorted w.r.t postsynaptic spike timing. The membrane potential w.r.t the spike timing of the postsynaptic neurons was also recorded. The top plot of Fig.~\ref{fig:vmem_stdp} shows the distribution of membrane potential approaching spiking threshold just before a postsynaptic spike, and is close to resting potential right after. The red curve on the bottom plot shows positive weight updates (LTP) when a presynaptic spike arrives just before the postsynaptic neuron fires and negative (LTD) when a presynaptic spike arrives just after the postsynaptic neuron fires. Beyond a certain time window, the mean weight update is zero, although individual updates have a large variance.

As can be seen in Fig.~\ref{fig:vmem_stdp}, the membrane based learning rule proposed here empirically approximates the pairwise exponential \ac{STDP} rule.


\subsection{Coincidence detection}
\label{subsec:coincidence}
\begin{figure}
\centering
\includegraphics[width=0.5\textwidth]{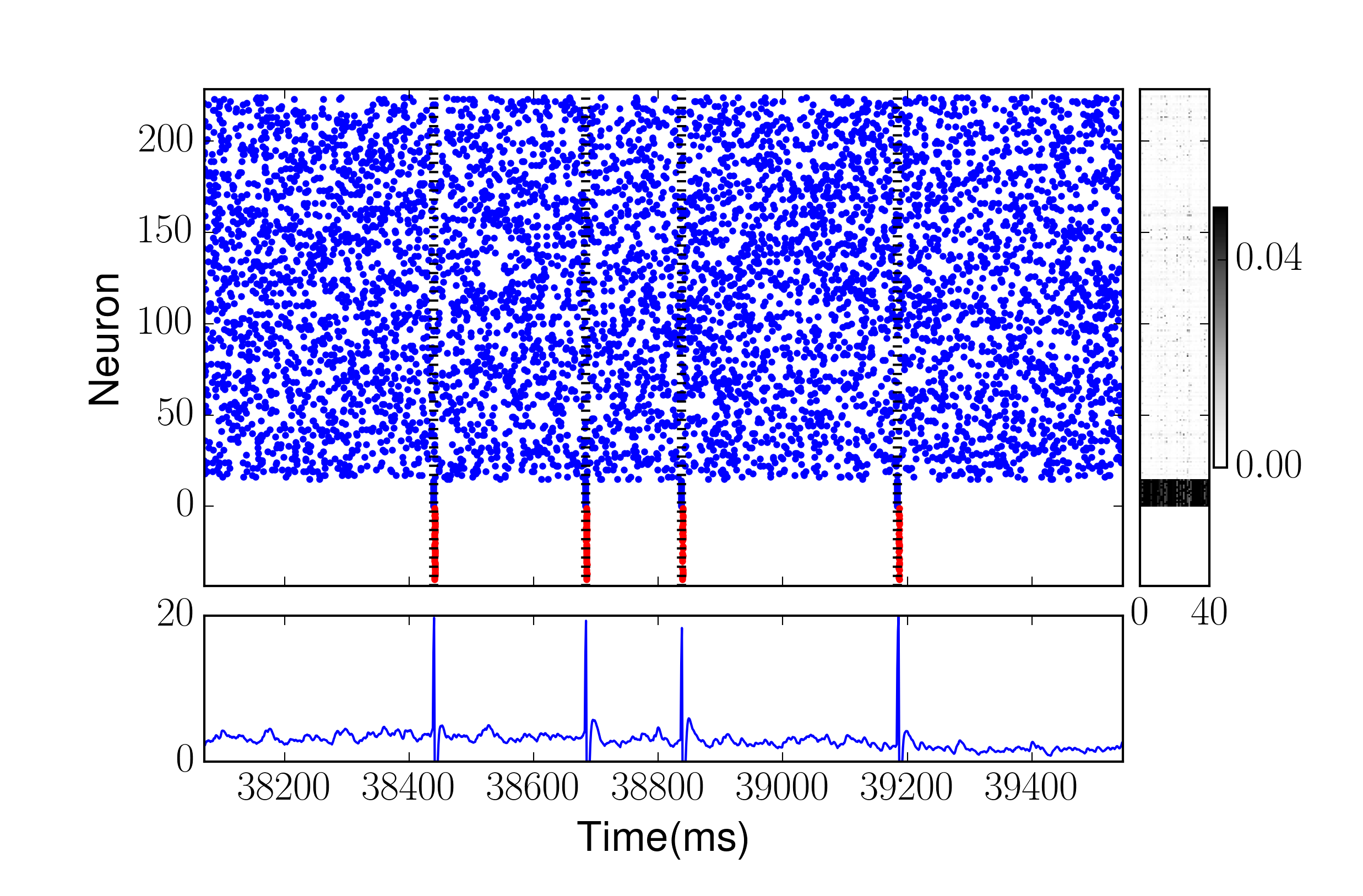}
\caption{Raster plot of spikes from $225$ inputs to $40$ neurons (in blue). The first $20$ of inputs are synchronous. The synaptic learning rule drives the weights of synchronously firing inputs to potentiate and other synaptic weights are depressed as can be seen in the grey scale image of the right sub-plot. As a result the post synaptic neurons only fire (in red) when there is coincidence of spikes. An example membrane potential of one of the $40$ neurons is shown in the lower sub plot.}
\label{fig:coincidence}
\end{figure}

A fundamental computational capability of \ac{STDP} empowered neurons is believed to be able to learn correlated activity in a form of Hebbian learning. To verify this ability with the proposed learning rule we setup the following experiment.
$40$ neuron receives inputs from $225$ spike sources over plastic synapses. 
$20$ of the inputs are generated from a single Poisson process with mean firing frequency of $5\,Hz$ and consequently the spikes from these inputs are always synchronous. The remaining input are generated from independent Poisson processes with mean spiking frequency of $20\,Hz$ (See Fig.~\ref{fig:coincidence}), and therefore lack synchronicity.

The weights of all $225$ synapses are initialized randomly from a uniform distribution. Over time the weight of synapses with synchronous inputs, \emph{ie.} coincident spiking activity drift to high weights. The input weights of synapses that lack synchrony are driven to low weights. As a result, the post synaptic neurons learn to fire only when there is coincident spiking activity from the first $20$ inputs.

\subsection{Hidden pattern detection}
\begin{figure}
\centering
\includegraphics[width=0.5\textwidth]{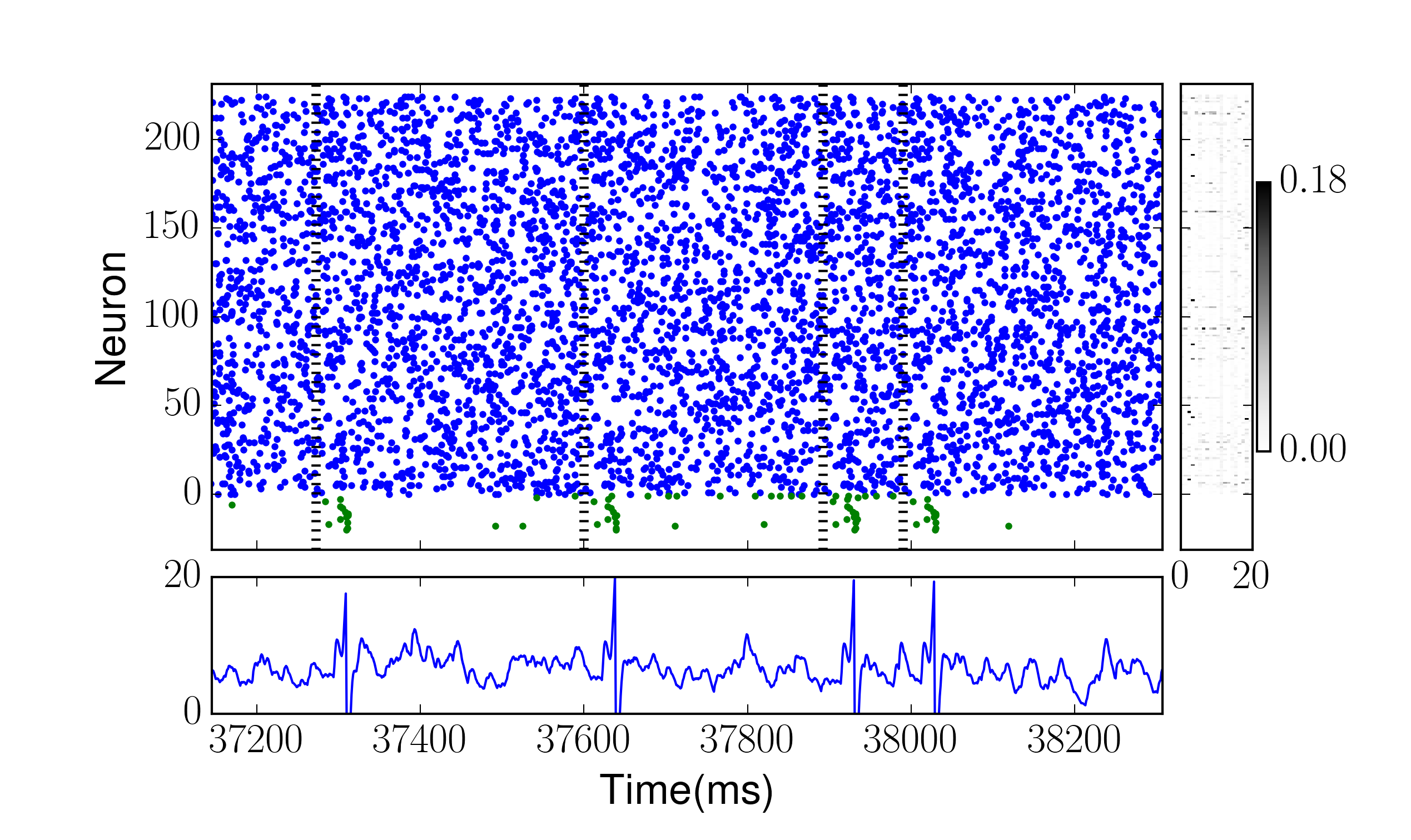}
\caption{Neurons receive Possion spike trains from $225$ pre-synaptic neurons. A $40\,ms$ spike pattern (SP) is repeatedly presented with a mean frequency of $5\,Hz$ (onset marked by vertical dotted lines) interspersed with random Poisson spikes of varying lengths. Both the spike pattern (SP) and random spikes have a mean spiking frequency of $20\,Hz$. After training with this input, the output neurons' response synchronizes with the presentation of SP as can be seen in the lower subplot. The final weights are shown in the right sub-plot.}
\label{fig:embedded}
\end{figure}

In the above subsection we presented inputs where some of the input sources always firing synchronously. We saw that under these circumstances the learning rule is able to potentiate the corresponding synapses and depress all other synapses. We now explore a different scenario when a randomly generated fixed spike pattern `SP' is repeatedly presented, interspersed with random spike patterns of the same firing rate. Fig.~\ref{fig:embedded} shows the presentation of such a stream of spikes to $20$ different neurons over $225$ synapses each. The initial weights were randomly initialized from a uniform distribution. Over time the synaptic weights converge such that the postsynaptic neurons selectively spike only on presentation of SP. It should be noted that not all neurons spike at the exact same time but fire in the neighborhood of SP, as can also be explained from the different final weights.

\subsection{Multiple pattern detection}
\begin{figure}
\centering
\includegraphics[width=0.5\textwidth]{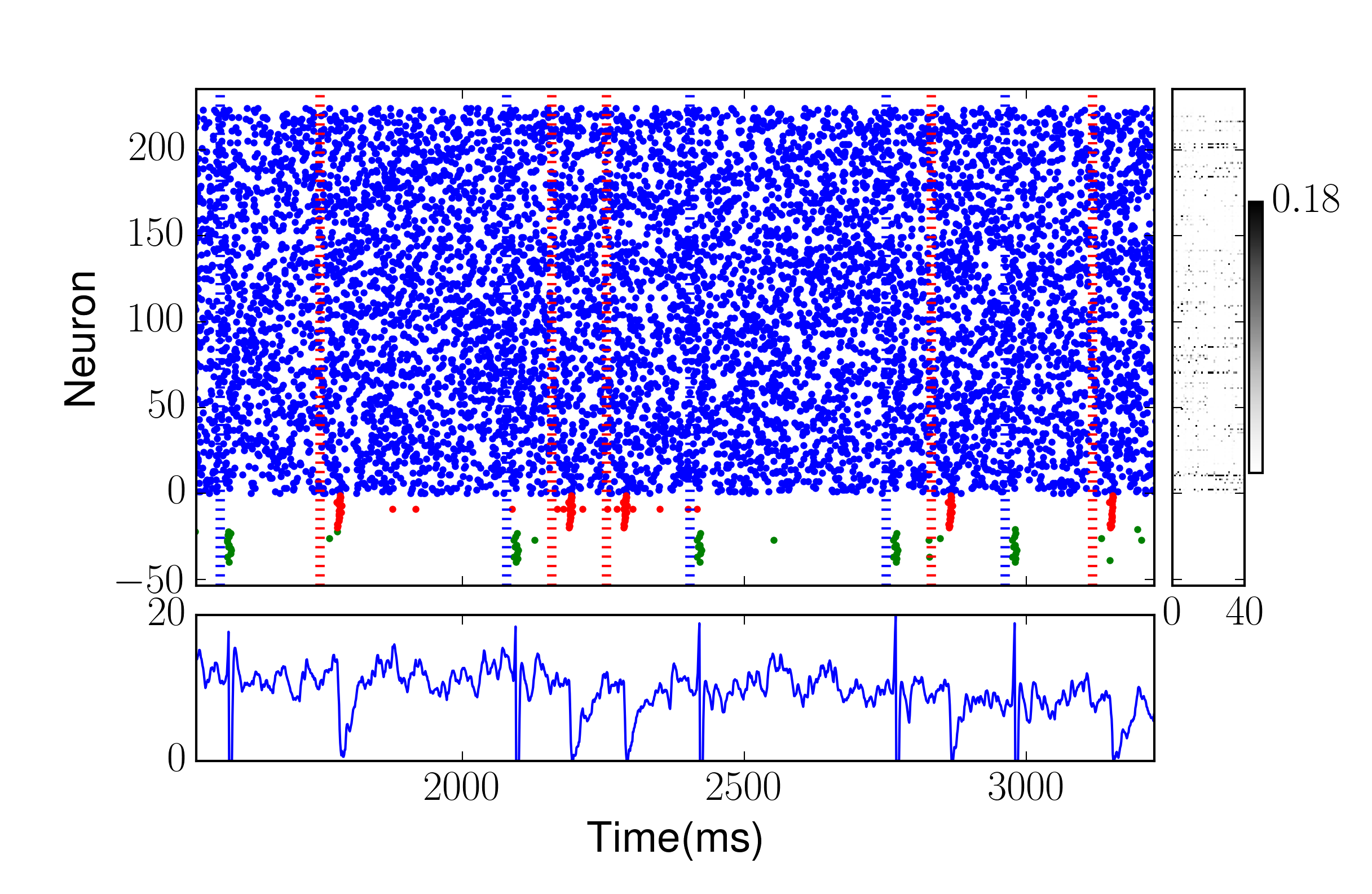}
\caption{Raster plot of input (blue) and output spike trains of two competing populations of neurons (red and green) learn to identify two different spike patterns (onsets marked with vertical dotted lines) embedded in noisy spike train.}
\label{fig:multiple_wta}
\end{figure}

The results above demonstrate that single patterns embedded in noise are detected and learned based on the learning rule. We now explore a more practical scenario where multiple patterns with statistical significance are presented to the neurons. More specifically, here we present two different patterns in random order interspersed with noise similar to the single pattern presentation experiment.
In order to ensure each of the coincidence detectors are tuned to one of the patterns of interest, a competitive winner-take-all is imposed on the coincidence detector neurons~\cite{Masquelier_etal2009,Habenschuss_etal2012}. Two populations of $20$ neurons each with self-excitatory connections and mutual inhibitory connections constitute this winner-take-all network of coincidence detectors. As shown in Fig.~\ref{fig:multiple_wta} the two populations indeed self organize to detect one of the two spike patterns. The competition allows the populations to converge onto different spike patterns.

\section{Discussion and Conclusion}

We propose a simple membrane based learning rule that enables individual neurons to learn and detect spatio-temporal spike patterns. In contrast to  other works that shown that membrane based learning rules can be used to perform supervised learning, here we demonstrate that such a rule can imbue unsupervised temporal learning capabilities to neurons.
We argue that, by only triggering weight updates from the pre-synaptic neuron, this learning rule is more amenable to hardware implementations where the weight is store using index based memory structures. This allows design of learning neuromorphic devices with minimal memory resources.

While this paper considers a single threshold on the membrane for the learning rule, future work is geared towards complex learning kernels on the membrane potential to mimic other \ac{STDP} kernels~\cite{Feldman_2012} and optimize the learning process.


%



%

\bibliography{biblio.bib}
\bibliographystyle{ieeetr}

\end{document}